\documentclass[letterpaper, 10 pt, conference]{ieeeconf}  

\IEEEoverridecommandlockouts                              

\usepackage{times}
\usepackage{graphicx}
\DeclareGraphicsExtensions{.pdf,.png,.jpg}
\usepackage{amsmath}
\usepackage{amssymb}  
\usepackage{makecell}
\usepackage{booktabs}
\usepackage{multirow}
\usepackage{here}
\usepackage{kotex}
\usepackage{soul, color}
\usepackage{caption}
\usepackage{subcaption}
\usepackage{changepage}
\usepackage{hyperref}

\title{\LARGE \bf
 Necessity Feature Correspondence Estimation for
 Large-scale
 Global Place Recognition and Relocalization
}

\author{Kyeongsu Kang$^{}$, Minjae Lee$^{}$ and Hyeonwoo Yu$^{}$
 \thanks{Kyeongsu Kang, Minjae   Lee and Hyeonwoo Yu are with the Department of Electrical Engineering \& Graduate School of AI, Ulsan National Institute of Science and Technology (UNIST), Ulsan, South Korea. {\tt\small \{thithin, lmjbsj, hyeonwoo.yu\}@unist.ac.kr}}%
 \thanks{The opensource implementation will be available at: \href{https://github.com/Lab-of-AI-and-Robotics/NFC_relocalization}{https://github.com/Lab-of-AI-and-Robotics/NFC\_relocalization} }
}

\begin{document}
	
\maketitle


\begin{abstract}


In order to find the accurate global 6-DoF transform by feature matching approach, various end-to-end architectures have been proposed.
However, existing methods do not consider the false correspondence of the features, thereby unnecessary features are also involved in global place recognition and relocalization.
In this paper, we introduce a robust correspondence estimation method by removing unnecessary features and highlighting necessary features simultaneously.
To focus on the necessary features and ignore the unnecessary ones, we use the geometric correlation between two scenes represented in the 3D LiDAR point clouds.
We introduce the correspondence auxiliary loss that finds key correlations based on the point align algorithm and enables end-to-end training of the proposed networks with robust correspondence estimation.
Since the ground with many plane patches acts as an outlier during correspondence estimation, we also propose a preprocessing step to consider negative correspondence by Weakening the influence dominant plane patches.
The evaluation results on the dynamic urban driving dataset, show that our proposed method can improve the performances of both global place recognition and relocalization tasks.
We show that estimating the robust feature correspondence is one of the important factors in global place recognition and relocalization.

\end{abstract}

\section{Introduction}

In the fields of the robotics and computer vision, wide-ranging researches have been conducted on Simultaneous Localization and Mapping (SLAM) \cite{DBLP:books/daglib/0014221}.
SLAM is widely used not only for mobile robots but also for unmanned technologies such as autonomous vehicles and underwater marine robots \cite{yu2019variational, jang2021multi, mittal2022vision}.
In traditional approaches, the pipeline of SLAM can be broadly divided into three parts: 1) front-end, 2) back-end, and 3) loop-closing \cite{cattaneo2022lcdnet}.
Loop closing can optimize the robot's pose when the robot detects a loop, also known as a Rendezvous, that connects to a previously visited location.
It is important for the robot navigation task to accurately detect loop closing point, because failure to detect loops can result in incorrect pose optimization of the map and the robot, and lead the distorted map representation. \cite{durrant2006simultaneous, bailey2006simultaneous}.
By successfully detecting previously visited locations, the robot's poses can be corrected through relocalization even if the robot fails to estimate its past trajectories correctly due to accumulation of errors.

\begin{figure}[t]%
    \centering%
    \includegraphics[scale=0.36]{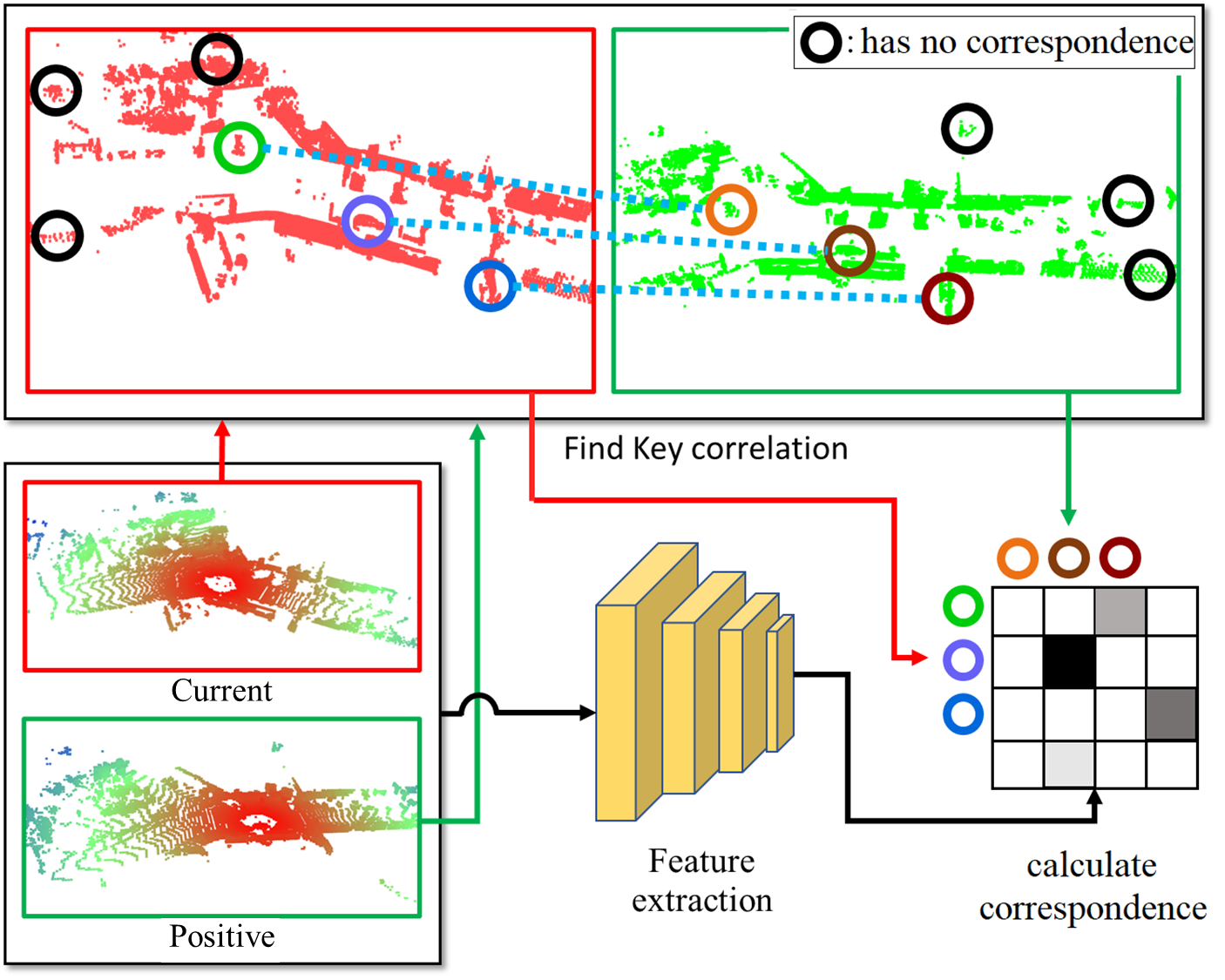}%
    \caption{%
          Our method enhances the performance of global place recognition and relocalization by making correspondence estimation robust. Robust correspondence estimation can improve the performance in loop closing and relocalization.
    }%
    \label{overview_test}%
    \vspace{-15pt}
\end{figure}

Place recognition is of significant importance for the loop closing detection, thereby many researches have been conducted.
However, there are still difficulties in performing place Recognition for 3D LiDAR point clouds, because 3D LiDAR point clouds are sparse compared to 2D images and have complex distributions of 3D geometric structures \cite{vidanapathirana2022logg3d}.
In this case, information that can distinguish geometric structures may not be incorporated into the features, which can cause the failure to detect course-level previously visited places.
It is also hard to achieve the fine-level pose estimation as well, because the estimation of the 6-DOF pose needs features that contain detailed structural information in 3D LiDAR point clouds.

To resolve these issues, in previous works, various registration methods based on ICP \cite{segal2009generalized} have been widely used to estimate a fine pose after coarsely recognizing the global pose \cite{dellenbach2022ct, shan2018lego, chen2021overlapnet, komorowski2021egonn}.
However, methods like ICP can be influenced by the initial pose and can fall into a local minimum, resulting in misaligned 3D scans.
Recent researches tackle these problems by estimating global place recognition and learning local descriptors for 6-DOF transform simultaneously or estimating initial pose for fine pose estimation \cite{chen2021overlapnet, du2020dh3d}.
To accomplish this, it is crucial to establish point correspondences, leading to the adoption of various methods, including data augmentation \cite{du2020dh3d} and unbalanced optimal transport \cite{cattaneo2022lcdnet}.
Nevertheless, these methods do not consider the necessity of feature points for estimating true-positive correspondence and simply utilize all possible points.
Using all feature points equally can result in incorrect correspondence estimation such as false-positive or false-negative one, as it tries to find correlations even if some of the feature points have no correspondence.
This leads to a lack of discrimination in local feature learning for structural information. 
As a result, when using coarse place recognition, unnecessary features are aggregated, leading to limited accuracy in global place recognition. 
Furthermore, the inability of estimating feature correspondence also results in poorer relocalization performance.


To overcome these limitation, we propose a method considering the necessity of each feature point for both global place recognition and relocalization.
By considering the necessity of each feature point based on the geometric characteristics before training, the network can estimate the key correspondences for each feature.
Robust estimation of the correspondence can reduce the number of the outliers, thus make it easier to find correspondences at each local point and leading to better results in 3D point cloud registration.
With fewer outliers, the accuracy of 6-DOF calculations during RANSAC or SVD will increase, enabling robust point cloud registration \cite{li2019usip}.
In addition, the better ``local consistency" of features can also lead to improvement in global place recognition \cite{vidanapathirana2022logg3d}.

To determine the necessity of each feature point for learning, this paper employs two techniques that consider geometrical feature correlation.
First, we weaken the influence of ground structures that have unnecessary geometrical features.
The features on the ground can be considered outliers from a local feature perspective because they have structurally duplicated forms, which can cause inaccurate correspondence information.
By weakening the influence of dominant ground we can remove features that can generate inaccurate correspondence.
Second, we exploit the weights from probabilistic model based on geometric correlation to determine the necessity of the features. 
Since the feature points we want to find correspondences in two scenes have a geometric correlation, we found that it is possible to find and learn the strong necessity of inlier features by considering this geometric correlation.
After finding the weights, the points corresponding to those weights are used to estimate correspondence according to the weights.


To summarize, we propose a method that enhances global place recognition and point cloud registration, by considering the geometric characteristics of 3D point clouds.
We weaken unnecessary feature points and strengthen necessary ones while training by feature correspondence estimation.
Our contribution is as follows.
\begin{itemize}
    \item 
    We preprocess the ground to weaken unnecessary features identified through geometrical character.
    The ground has repetitive structures due to its characteristics, and these characteristics can become irrelevant in feature extraction or clustering \cite{shan2018lego, jiang2020lipmatch, yang2019monocular}.
    Therefore, ground can act as outliers when estimating point correspondence between two scenes, so they are weakened through the preprocessing process.
    \item 
    We enhance learning features with better matching, considering probabilistic model based on geometric correlation through 3D alignment algorithms.
    Each overlapping points have corresponding relationships. 
    We estimate correspondence using weights computed by the probabilistic model based on geometric correlation.
\end{itemize}
    Our method shows better performance compared to the previous methods.
    We confirmed the high performance of our method through testing on the KITTI and KITTI-360 datasets. 
    This demonstrates the critical role of considering probability-based geometric correlation in global place recognition and relocalization.

\section{related works}
Research in place recognition in 3D point clouds can be divided into two main categories: global place recognition and point cloud registration \cite{kim20191}.
Global place recognition is a coarse-level place recognition method that utilizes feature points of a scene to search topological locations on a large-scale map.
Point cloud registration, on the other hand, is a fine place recognition method that calculates the 6-DOF of a scene to describe transformation in a metrical location perspective.
Although accurate transformations within a scene can be described with point cloud registration, it is often used as a supplementary method in large-scale place recognition due to limitations in computing feature correspondences.

Global place recognition uses a method that compares the global descriptor of the current point clouds to that of the map to determine their similarity.
The traditional approach is handcrafted-based method using BEV, generating a global descriptor \cite{kim2018scan, kim2021scan}.
Recently, an end-to-end learning-based global place recognition approach has been proposed, showing promising results \cite{uy2018pointnetvlad}.
This method combines point cloud feature extraction \cite{qi2017pointnet} with vision-based global place recognition \cite{arandjelovic2016netvlad}, demonstrating its viability for use in 3D point clouds.
More recently, architectures have been proposed to generate global descriptors that better represent the place, thereby improving the performance of global place recognition \cite{vidanapathirana2022logg3d, komorowski2021minkloc3d, Hui_2021_ICCV}.

Recent research has emerged with an architecture that combines point cloud registration and global place recognition \cite{chen2021overlapnet, komorowski2021egonn}. 
They aim for more accurate place recognition by not only learning global descriptors but also local features that can describe 6-DOF transforms. 
To learn local features, correspondence must be known, and methods have been proposed to augment training data to achieve \cite{du2020dh3d}. 
In the case of \cite{cattaneo2022lcdnet} used sparse convolution network \cite{shi2020pv} to extract features and proposed unbalanced optimal transport to find correspondence between features. 
Our work also effectively estimates correspondence through the use of unbalanced optimal transport.

\begin{figure*}[t]%
    \centering%
    \includegraphics[scale=0.52]{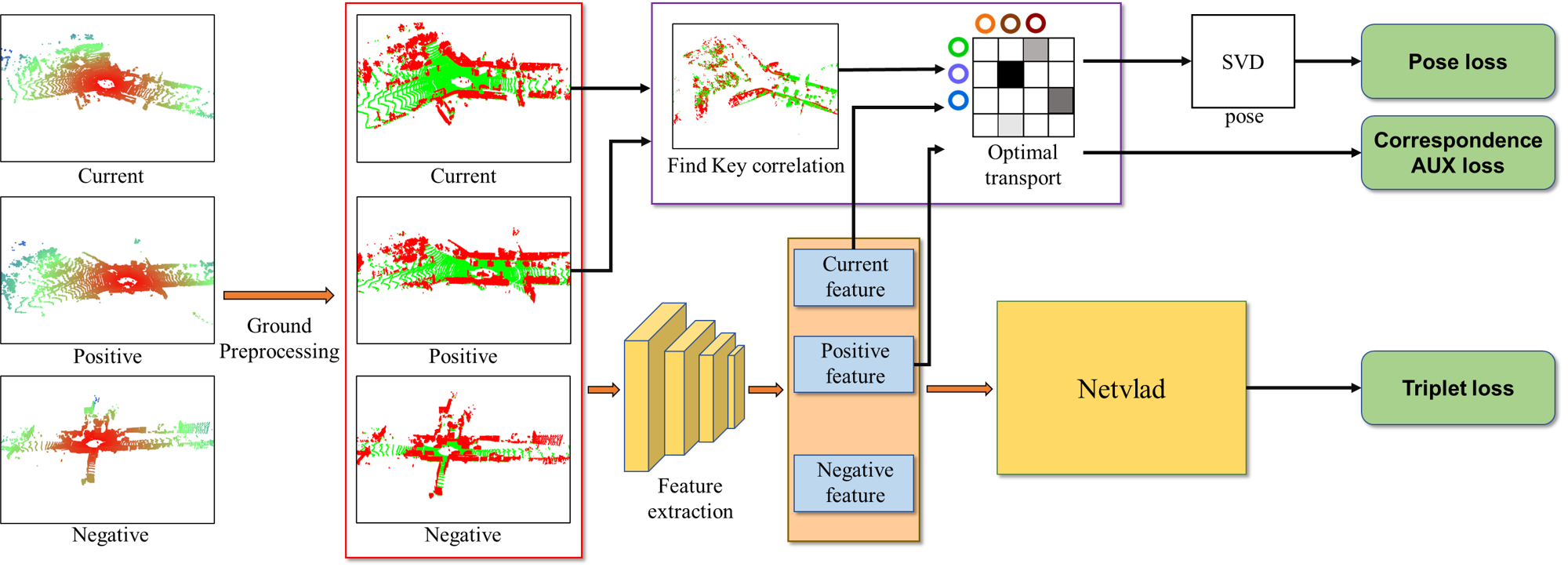}%
    \caption{%
    The overview of our proposed technique.
    The determined key correlations and features are used to judge correspondence, and ultimately, the training is done through pose loss, correspondence auxiliary loss, and triplet loss.
    }%
    \label{overview}%
    \vspace{-15pt}
\end{figure*}

\section{Approach}


Suppose we have a set of points $L=\{l_n\}$ and the corresponding local feature set $F=\{f_n\}$.
We aim to have the optimal feature set $\hat{F}$ which can be fully utilized both for global place recognition and relocalization.
Here, let $C$ be the ground truth correspondence of $L^A$ and $L^B$ detected in scenes $A$ and $B$ respectively.
In order to obtain $\hat{F}$, it is necessary to estimate $C$. 
However, as $C$ remains unknown in usual, estimating $C$ is the foremost thing that we need.
Assume we have $H$, which is a function that finds correspondences between two point sets.
Then $\tilde{C}$ that estimate $C$ of ${L^A}$ and ${L^B}$ is represented as the following:
\begin{align}
    \label{correspondenceAB}  
    \tilde{C}=H \left( \{ L^A,F^A \} , \{ L^B,F^B \} \right),
\end{align}
where $F^A$ and $F^B$ is local feature extracted from $L^A$ and $L^B$ respectively.
To solve this data association problem, unbalanced optimal transport can serve as one of the $H$ function for correspondence estimation \cite{cattaneo2022lcdnet, xu2023ring++, sejourne2022unbalanced}.
The $\tilde{C}$ derived through unbalanced optimal transport is presented as the following:

{\footnotesize 
\begin{align}  
\nonumber
    \tilde{C} & =  \underset{Z \in \mathbb{R}^{N \times N}}{\arg \min }\left\{\left(\sum_{i, j} K_{i j} Z_{i j}+\lambda Z_{i j}\left(\log Z_{i j}-1\right)\right)+\right. \\
     & \left.\rho\left(D_{KL} \left(\sum_i Z_{i j} || U(1, N)\right)+D_{KL} \left(\sum_j Z_{i j} || U(1, N)\right)\right)\right\}
     \label{optimalT}
    ,
\end{align}
}

\noindent
where $D_{KL}$ is the Kullback–Leibler divergence, $U$ is the discrete uniform distribution, $\rho$ is a parameter that mass is preserved and $\lambda$ is a parameter that the sparseness of the mapping.
$K_{i j}$ is a dependent variable on $F^{A}$ and $F^{B}$ which represents the matching cost of the $i$'th point in $L^{A}$ and $j$'th point in $L^{B}$. 

Using Eqn.~\eqref{optimalT}, existing methods obtain $\tilde{C}$ and estimate optimal local features subsequently.
However, when estimating $\tilde{C}$ through unbalanced optimal transport, it is challenging to consider prior information such as geometric characteristics of $L$, because the existing approach simply applies uniform weights $U$ to every possible correspondenses of $L^{A}$ and $L^{B}$. This limitation makes it hard to consider true positive and false positive, which is crucial for relocalization, and furthur global place recognition as well.

Therefore, we propose a method that enhances both global place recognition and relocalization performance by strengthening the true-positive correspondences and weakening the False-positive ones by considering geometric correlation.
By using robust correspondence estimation based on the necessity of feature points, it is easier to find the feature association for local points.
Our approach can lead to fewer outliers and improved accuracy in 6-DOF when calculating RANSAC or SVD, reducing the error in metric pose and relocalization.
Additionally, as the local feature correspondence was improved, ``local consistency" can be maintained, leading to an improvement in the performance of global place recognition \cite{vidanapathirana2022logg3d}.

The overview of our proposed method is displayed in Fig.~\ref{overview}. 
First, when the current points are provided, we consider areas with repetitive and unnecessary geometrical features as the ground and preprocess those sections.
Then, local features are extracted from the preprocessed data using the local feature extraction module.
At the same time, overlapping points distance are found by performing ICP or a better align algorithm on the preprocessed data between the current point cloud and positive.
The overlapping points found through the local features and the align algorithm are then estimated through weight unbalanced optimal transport for correspondence, and the pose is obtained through SVD.
The obtained pose is then trained through correspondence auxiliary loss and pose loss.
Additionally, the local features are used to generate a global descriptor through the aggregation module, NetVLAD \cite{arandjelovic2016netvlad}.
The generated global descriptor can be trained through triplet loss.

\subsection{Weakening the Unnecessary Correspondences}

Because of the structural characteristics of ground, preprocessing of ground has a significant impact on the performances. 
Previous researches on global place recognition showed that removing ground and training results in better performance, or preprocessing ground before training is necessary \cite{komorowski2021minkloc3d, dube2018segmap, zhang2019pcan}.
However, they hardly consider the effectiveness of ground preprocessing in the aspect of the feature correspondence for global place recognition and relocalization.

Local features in 3D LiDAR place recognition are created by condensing the geometric shape information inherent in the point cloud.
In other words, local features depend on the geometric information contained within the point cloud \cite{qi2017pointnet, NIPS2017_d8bf84be}.
In 3D LiDAR data, the ground exhibits a continuous plane structure, resulting in the inclusion of numerous repetitive plane structure details in the local features during feature extraction \cite{shan2018lego}.
These repetitive plane structure details can act as outliers that lead to incorrect geometric correlations between two scenes, causing confusion in feature correspondence.
Therefore, data with a continuous plane structure, such as the ground, can hinder correspondence during the model training process and potentially degrade performance.
By weakening the influence of structures with repetitive plane patterns from a geometric perspective, we find better correspondence between local features in place recognition.
We propose a method to mitigate the impact of unnecessary correspondence from this perspective.
Geometric characteristics that are indicative of repetitive flat plane structures include flatness, uprightness, elevation, and laneness \cite{lim2021patchwork}. 
We leverage these attributes to assess the ground using a probabilistic model.
The ground can be determined by the following:
\begin{align}
    \label{ground_removal} 
    P_G(L_k) = \prod_i P_{Gi}\left(L_k\right)
    ,
\end{align}           
where $L_k$ is the $k$'th section in LiDAR point cloud, $P_{Gi}(L_k)$ is $k$'th section point probability that the $k$'th section of LiDAR point from the perspective of planeness, uprightness, elevation, and flatness is ground.
$P_G(L_k)$ is the probability that the $k$'th section of LiDAR point is ground, expressed as the product of $P_{Gi}(L_k)$.
If $P_G(L_k)$ in Eqn.~\eqref{ground_removal} is high enough, the $k$'th section is estimation ground.
Detailed formulas are explained in \cite{lim2021patchwork}.
By weakening the influence of the ground that may induce false-positive correspondece, outliers are eliminated by having lower probability when estimating correspondences, enabling more accurate feature association.

\subsection{Strengthening the Necessary Correspondences}

We also consider the geometric correlation between points in two scenes to assess the importance of features for correspondence estimation.
By considering the geometric correlation between points in two scenes, robust correspondence estimation can be made, leading to a reduction in outliers and improvement in performance of place recognition.
This results in improved estimation of relocalization and global place recognition. 

To enhance the correspondence accuracy based on feature importance assessment, we rewrite Eqn.~\ref{correspondenceAB} as the following:
\begin{align}
    \tilde{C} = H \left(\{L^{A}, F^{A}\}, \{L^{B}, F^{B} \}, W \right)
    ,
    \label{modified_H}
\end{align}
where $W$ is the geometric correlation weight for $L^{A}$ and $L^{B}$.
This weight matrix $W$ is defined as:
\begin{align}
\nonumber
    W_{i j}= P_{i j}(1-P_G(L^A_i))(1-P_G(L^B_i)) ,
\end{align}
where
$P_G(L^A_i)$ and $P_G(L^B_i)$ are probabilistic models of the ground for $L^A_i$ and $L^B_i$ in Eqn.~\eqref{ground_removal}. $P_{i j}$ is defined as:
\begin{align}
\label{weights} 
    P_{i j} = e^{-D(L^A_i, L^B_j)}.  
\end{align}
Here, $P_{i j}$ represents the probabilistic model of geometric relationship, where a stronger geometric relationship is assumed when the geometric distance $D(L^A_i, L^B_j)$ between $L^A_i$ and $L^B_i$ is closer.
Therefore, matrix $W$ determines whether two scenes in the LiDAR point cloud have correspondences with each other.
In Eqn.~\eqref{weights}, as the geometric distance between $L^A_i$ and $L^B_j$ decreases, signifying a tighter geometric connection, we allocate a greater weight $W_{ij}$ to establish this correspondence.

Since we rewrite $H$ using $W$ as in Eqn.~\eqref{modified_H}, our estimated correspondence $\Tilde{C}^\prime$ can be given as the following:
\vspace{-1pt}
{\footnotesize 
\begin{align}
\nonumber
    & \tilde{C}^\prime =  \underset{Z \in \mathbb{R}^{N \times N}}{\arg \min }\left\{\left(\sum_{i, j} K_{i j} W_{i j}Z_{i j}+\lambda W_{i j}Z_{i j}\left(\log W_{i j}Z_{i j}-1\right)\right) \right. +\\
     & \left.\rho \left(K L\left(\sum_i W_{i j}Z_{i j} \mid U(1, N)\right)+K L\left(\sum_j W_{i j}Z_{i j} \mid U(1, N)\right)\right)\right\}
    .
    \label{modified_optimalT}
\end{align}
}

\noindent
However, we found that when we directly apply Eqn.~\eqref{modified_optimalT}, an issue arises where points with lower weights risk getting truncated, thereby missing the opportunity to be considered for the optimal correspondence. Therefore, we directly multiply $W$ to $\tilde{C}$ in order to obtain $\tilde{C}^\prime$ in practice: $\tilde{C}^\prime=W\tilde{C}$.
%

Using $W$ and $\tilde{C}^\prime$, we finally have the optimal feature $\hat{F}$ as the following \cite{cattaneo2022lcdnet}:
\begin{align}
    \label{correspondenceloss}
    \hat{K}
    =
    \underset{K}{\arg \min }
    \text{        }
    \mathbb{E}\left[\left|\frac{\sum_{j=1}^N \tilde{C}^\prime_{i j} L_j^{B}}{\sum_{j=1}^N \tilde{C}^\prime_{i j}}-T_A^B W_{i j} L_i^{A}\right|\right]
    ,
\end{align}
where $T_A^B$ is ground truth transformation from B to A.
Since $K$ is the matching cost determined by $F_A$ and $F_B$, we can have $\hat{F}$ from Eqn.~\eqref{correspondenceloss}. 
Since $K$ and $\tilde{C}^\prime$ are correlated to each other in Eqn.~\eqref{modified_optimalT} and Eqn.~\eqref{correspondenceloss}, we iteratively perform optimizations for both $K$ and $\tilde{C}^\prime$, similar to Expectation-maximization manner.

The proposed method, which takes into account geometric correlations, enhances the learning process by placing greater emphasis on points that are likely to have a genuine correspondence $C$.

\begin{table*}[!htbp]
\setlength{\tabcolsep}{4pt}
\renewcommand{\arraystretch}{1.1}
    \caption{\\ \small GLOBAL PLACE RECOGNITION PERFORMANCE ACCORDING TO EACH METHOD}
\vspace{-10pt}
    \label{globaltable}
    \begin{center}%
        \begin{tabular}{@{\extracolsep{4pt}}l|| cccc c cccc @{}}
            \hline

            \hline
            \multirow{3}{*}{Method}  & \multicolumn{4}{c}{AP Metrics 1}  & & \multicolumn{4}{c}{AP Metrics 2 } \\
            \cline{2-5} \cline{7-10} 
            & \multicolumn{2}{c}{KITTI} & \multicolumn{2}{c}{KITTI-360} & &  \multicolumn{2}{c}{KITTI} & \multicolumn{2}{c}{KITTI-360}\\
            \cline{2-3} \cline{4-5} \cline{7-8} \cline{9-10}
            & 00 & 08 & 02 & 09 & & 00 & 08 & 02 & 09 \\
            \hline

            \hline
            Scan Context \cite{kim2018scan} & 0.96 &  0.65 &  0.81 & 0.90 & & 0.47 & 0.21 & 0.32 & 0.31 \\
            OverlapNet \cite{chen2021overlapnet}&  0.95 & 0.32 & 0.14 & 0.70 & & 0.60 & 0.20 & 0.05 & 0.33 \\
            LCDNet \cite{cattaneo2022lcdnet} & 0.97 & 0.88 & 0.87 & 0.91 & & 0.95 & 0.73 & 0.76 & 0.83 \\
            Ours & \textbf{0.99} & \textbf{0.90} & \textbf{0.91} & \textbf{0.94} & & \textbf{0.98} & \textbf{0.79} & \textbf{0.82} & \textbf{0.88} \\
            \hline
            
        \hline
        \end{tabular}
    \end{center}%
\vspace{-15pt}
\end{table*}

\begin{table*}
\setlength{\tabcolsep}{4pt}
\renewcommand{\arraystretch}{1.1}
    \caption{\\ \small POINT CLOUD REGISTRATION PERFORMANCE ACCORDING TO EACH METHOD}
\vspace{-10pt}
    \label{localtable}
    \begin{center}%
        \begin{tabular}{@{\extracolsep{4pt}}l|| cccc c cccc c cccc @{}}
            \hline

            \hline
            \multirow{3}{*}{Method}  & \multicolumn{4}{c}{Success rate [\%]}  & & \multicolumn{4}{c}{RME [DEG]} & & \multicolumn{4}{c}{TME [m]} \\
            \cline{2-5} \cline{7-10} \cline{12-15}
            & \multicolumn{2}{c}{KITTI} & \multicolumn{2}{c}{KITTI-360} & &  \multicolumn{2}{c}{KITTI} & \multicolumn{2}{c}{KITTI-360} & &  \multicolumn{2}{c}{KITTI} & \multicolumn{2}{c}{KITTI-360}\\
            \cline{2-3} \cline{4-5} \cline{7-8} \cline{9-10} \cline{12-13} \cline{14-15}
            & 00 & 08 & 02 & 09 & & 00 & 08 & 02 & 09 & & 00 & 08 & 02 & 09 \\
            \hline

            \hline
            Scan Context \cite{kim2018scan} & - &  - &   - & - & & 1.92 & 3.11 & 5.49 & 6.80 & & - & - & - & - \\
            OverlapNet \cite{chen2021overlapnet} &  - &  - & - & - & & 3.89 & 65.45 & 76.74 & 33.62 & & - & - & - & -  \\
            LCDNet \cite{cattaneo2022lcdnet} &  {93.14} & \textbf{61.75} &  {82.23} &  {90.61} & &  {0.88} &  {2.73} &  {1.61} &  {1.08} & &  {0.77} & \textbf{1.65} &  {1.16} &  {0.91} \\
            Ours & \textbf{96.39} &  {59.59} & \textbf{89.84} & \textbf{93.99} & & \textbf{0.70} & \textbf{2.62} & \textbf{1.38} & \textbf{0.92} & & \textbf{0.59} & {2.13} & \textbf{0.89} & \textbf{0.73}   \\
            \hline
            
        \hline
        \end{tabular}
    \end{center}%
\vspace{-15pt}
\end{table*}

\begin{table*}[t]%
	\caption{\\ \small COMPARISON OF PERFORMANCE AMONG DNN-BASED METHODS}%
\renewcommand{\arraystretch}{1.1}
\vspace{-10pt}
	\label{DDNbased}%
	\begin{center}%
		\begin{tabular}{c|c|c||@{\extracolsep{4pt}}ccc  ccc @{}}
                \hline
   
                \hline
			& \multirow{2}{*}{Dataset} & \multirow{2}{*}{Sequences} & \multicolumn{3}{c}{Global Place Recognition} & \multicolumn{3}{c}{Point Cloud Registration} \\
            \cline{4-6} \cline{7-9}
			&  &  & Max F1 & recall@1 [\%] & recall@5 [\%] & successful rate [\%] & RME [DEG] & TME [m] \\
            \hline
                
                \hline
                LCDNet & \multirow{6}{*}{KITTI} & \multirow{2}{*}{00} & 0.927 & 97.681 & 98.870 & 93.14 & 0.876 & 0.775  \\             
                Ours   &                        &  & \textbf{0.951} & \textbf{98.870} & \textbf{99.762} & \textbf{96.39} & \textbf{0.703} & \textbf{0.597}  \\
                \cline{0-0} \cline{3-9} 
                LCDNet &  & \multirow{2}{*}{02} & 0.847 & 88.317 & 94.853 & 76.21 & 0.989 & 1.432 \\             
                Ours   &  &  & \textbf{0.899} & \textbf{93.045} & \textbf{99.304} & \textbf{82.32} & \textbf{0.956} & \textbf{1.198}   \\
                \cline{0-0} \cline{3-9} 
                LCDNet &  & \multirow{2}{*}{08} & 0.812 & 84.250 & 96.941 & \textbf{61.75} & 2.730 & \textbf{1.645}\\             
                Ours   &  &  & \textbf{0.836} & \textbf{87.461} & \textbf{97.094} & 59.59 & \textbf{2.624} & 2.130  \\
                \hline 

                \hline
                LCDNet & \multirow{10}{*}{KITTI-360} & \multirow{2}{*}{02} & 0.798 & 88.678 & 95.809 & 82.23 & 1.606 & 1.164\\             
                Ours   &                        &  & \textbf{0.831} & \textbf{92.520} & \textbf{98.282} & \textbf{89.83} & \textbf{1.380} & \textbf{0.891}  \\
                \cline{0-0} \cline{3-9} 
                LCDNet &  & \multirow{2}{*}{04} & 0.783 & 89.675 & 95.838 & 87.72 & 1.397 & 0.996\\             
                Ours   &  &  & \textbf{0.793} & \textbf{93.257} & \textbf{97.952} & \textbf{92.59} & \textbf{1.178} & \textbf{0.789}  \\
                \cline{0-0} \cline{3-9} 
                LCDNet &  & \multirow{2}{*}{05} & 0.811 & 85.647 & 93.442 & 86.03 & 1.626 & 1.034\\             
                Ours   &  &  & \textbf{0.816} & \textbf{89.288} & \textbf{96.476} & \textbf{93.07} & \textbf{1.248} & \textbf{0.778}    \\
                \cline{0-0} \cline{3-9} 
                LCDNet &  & \multirow{2}{*}{06} & 0.879 & 90.940 & 97.241 & 86.52 & 1.257 & 1.025\\             
                Ours   &  &  & \textbf{0.880} & \textbf{93.054} & \textbf{98.147} & \textbf{89.52} & \textbf{1.175} & \textbf{0.876}   \\
                \cline{0-0} \cline{3-9} 
                LCDNet &  & \multirow{2}{*}{09} & 0.828 & 91.047 & 96.654 & 90.61 & 1.084 & 0.905\\             
                Ours   &  &  & \textbf{0.871} & \textbf{94.722} & \textbf{98.645} & \textbf{93.99} & \textbf{0.927} & \textbf{0.732}   \\
                \hline
                
            \hline
		\end{tabular}%
	\end{center}%
    \vspace{-15pt}
\end{table*}%

\section{Experiments}

\subsection{Implementation and Setup}
To evaluate the performance of place recognition, we check the results from both the global place recognition and point cloud registration perspectives.
we use the KITTI odometry dataset \cite{geiger2012we} and KITTI-360 dataset \cite{Liao2022PAMI} to evaluate our proposed method.
The KITTI and KITTI-360 datasets are LiDAR datasets, making them suitable for evaluation due to their inclusion of moving vehicles in various dynamic urban environments.
KITTI Sequence 00 have same driving direction of the loop, and KITTI sequence 08 has inverse driving direction of the loop.
KITTI-360 sequences 02 and 09 both have driving loops in both same and inverse directions, and they contain the highest number of loop closures.
Therefore, We used KITTI sequences 00, 08 and KITTI-360 sequences 02, 09 for testing.
Hence, we employ these sequences to assess the performance of our method across loops originating from both forward and reverse driving directions.
In addition, for a detailed performance comparison with the state-of-the-art DNN-based method, LCDNet, we conducted tests on KITTI sequences 00, 02, 08 (excluding sequences 05, 06, 07, 09 that used to train) and KITTI-360 sequences 02, 04, 05, 06, and 09.

Similar to \cite{cattaneo2022lcdnet}, to train and evaluate our method we designate locations within 4 meters as positive (same place) pairs and locations more than 10 meters apart as negative (different place) pairs at the source point cloud and the target map.
For preprocessed ground training, we remove the ground that uses the probability model \cite{lee2022patchwork++} to estimate the ground and only use non-ground points for training.
In training using $\mathcal{L}_{C A U X}$ in Eqn.~\eqref{correspondenceloss}, we find the overlap points
on 4096 keypoints extracted from baseline in 3D coordinates. 
To emphasize the effect of geometric relationships, when the points of two scenes that have registered point clouds are within 0.5m of each other, it is considered a point of the overlapping location.
In the case of overlapping location, we assign a value of 0 to $D(L^A_i, L^B_j)$ in Eqn.~\eqref{weights}. Conversely, for points that do not overlap, we assign $D(L^A_i, L^B_j)$ with a value of $\infty$.
We train the DNN-based model \cite{cattaneo2022lcdnet} and Ours on KITTI sequences 05, 06, 07, and 09.
All methods use a batch size of 6 and epoch of 150 and are trained with the same hyperparameters on NVIDIA 4090 RTX GPU.

\subsection{Global Place Recognition}
We evaluate our proposed method from the perspective of global place recognition using two metrics: AP (Average Precision) Metric 1 and AP Metric 2.
AP Metric 1 evaluates loop-closing detection performance for the pair with the highest similarity between the current point cloud and the previous point cloud. 
In contrast, AP Metric 2 evaluates loop-closing detection performance for not only the pair with the highest similarity but also all previous point clouds.
In real-world loop detection, the measurement approach of AP Metric 1 can be more accurate.
However, to assess loop detection performance in challenging scenarios, AP Metric 2 may be a better evaluation metric \cite{cattaneo2022lcdnet}.
We evaluate the global place recognition performance by comparing our method with state-of-the-art techniques, including Scan Context \cite{kim2018scan}, OverlapNet \cite{chen2021overlapnet}, and LCDNet \cite{cattaneo2022lcdnet}.
The results can be found in Table~\ref{globaltable}.
Additionally, the global place recognition section in Table~\ref{DDNbased} provides a detailed comparison between the DNN-based methods, LCDNet and our method, by summarizing Max F1, recall@1, and recall@5, which represent the percentage of correct answers within the top 1\% and 5\% of candidates.
The metrics with the highest performance in Table~\ref{globaltable} and Table~\ref{DDNbased} are highlighted in bold symbol.

In most cases, the AP Metric 1 approach shows similar results for forward loop detection, but significant differences are observed in reverse loop detection.
Particularly, OverlapNet exhibits notably low performance in sequences with reverse loops, such as KITTI sequence 08 and KITTI-360 sequence 02, and Scan Context also drops to 0.65 in KITTI sequence 08.
In the case of AP Metric 2, both OverlapNet and Scan Context demonstrate low performance in KITTI sequence 08, which includes reverse loops, with an AP of 0.20 and 0.21.
On the other hand, Our method does not show a significant performance drop. 
At the same time, Our method consistently outperforms LCDNet across all sequences, achieving the highest performance. 
The performance comparison between LCDNet and Our method in Table~\ref{DDNbased} also shows higher performance for Our method across all sequences. 
This result demonstrates that considering the necessary information for correspondence estimation helps in rotation and translation invariant global place recognition.

\begin{figure}[!tb]
\begin{adjustwidth*}{-0cm}{-0cm}
        \centering
        \begin{subfigure}[b]{0.2\textwidth}
            \centering
            \includegraphics[width=\textwidth]{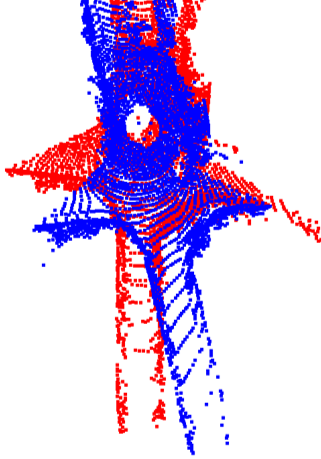}
            \caption[]%
            {{\small LCDnet}}    
            \label{}
        \end{subfigure}
        \hspace{20pt}
        \begin{subfigure}[b]{0.2\textwidth}  
            \centering 
            \includegraphics[width=\textwidth]{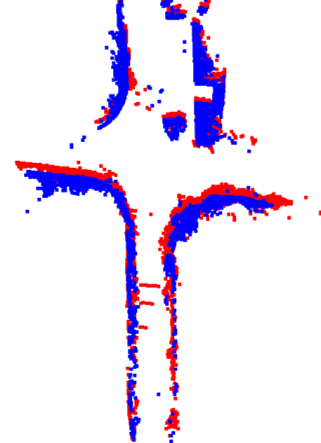}
            \caption[]%
            {{\small Ours}}    
            \label{}
        \end{subfigure}

        \caption[]
        {\small 
        Point Cloud Registration Results According to LCDNet and Our Method. LCDNet (a) exhibits errors in registration, whereas our method (b) successfully performed effective registration.
        }
        \label{comparemodel}
   \vspace{-15pt}
\end{adjustwidth*}
\end{figure}

\subsection{Relocalization}
We also evaluate our proposed technique from the point cloud registration perspective using Success rate, RME (Rotation Mean Error), and TME (Translation Mean Error).
Successful rate indicates the success ratio of the point cloud registration; where we consider a case to be successful if both the rotation error is less than 5 degrees and the translation error is less than 2 meters. 
Similar to the global place recognition evaluation, we assess point cloud registration performance by comparing it with state-of-the-art techniques, including Scan Context, OverlapNet, and LCDNet.
Both LCDNet and ours can calculate the 6-DOF transform between the two point clouds using UOT-based relative position estimation, while Scan Context and OverlapNet can only estimate the yaw angle between point clouds. 
Therefore, Scan Context and OverlapNet measure metrics related to RME only.
The results can be found in Table~\ref{localtable}.
The point cloud registration section in Table~\ref{DDNbased} provides a detailed comparison between the DNN-based LCDNet and our method, summarizing the performance for success rate, RME, and TME across multiple sequences.
The metrics with the highest performance in Table~\ref{localtable} and Table~\ref{DDNbased} are highlighted in bold symbol.

In the case of RME, all methods perform well in forward loops, but as with global place recognition, differences in performance are noticeable in reverse loops.
OverlapNet, in particular, shows a significant error of 76.74 degrees in KITTI-360 sequence 02, indicating its inability to estimate rotation accurately in reverse loops. 
Scan Context also exhibits RME measurements exceeding 5 degrees in some sequences.
However, DNN-based models consistently achieve RME measurements below 5 degrees for all sequences.
Additionally, Our method outperforms most performance metrics, not only in terms of RME.
LCDNet achieves higher performance in success rate and TME for KITTI sequence 08.
However, according to the summarized point cloud registration in Table 3, our method outperforms in all other sequences.
As shown in Fig.~\ref{comparemodel}, it can be observed that our method consistently achieves higher point cloud registration performance in most cases due to its robust feature correspondence.
This result highlights the importance of feature correspondence, as predicted by geometric correlation, in point cloud registration.

\section{Conclusion}
In this paper, we presented a method of robust feature correspondence estimation for improving 3D place recognition and relocalization. 
We proposed a probabilistic technique to enhance the necessary features and suppress the unnecessary features simultaneously in the estimation of correspondence, by considering the geometric characteristics of the 3D scenes.
To weakening the strong unnecessary correspondence, we first showed that features on the ground can act as outliers when estimating correspondence therefore should be preprocessed. 
We also found the key correlations through the geometric registration method to estimate robust correspondence by focusing on necessary features which can enhance both global place recognition and relocalization. 
We provided the evaluation results to show improvement performances in the place recognition problem on the KITTI odometry dataset and KITTI-360 dataset, an urban dynamic environment dataset. 
Our approach demonstrated significant improvement in both 3D point cloud registration and global place recognition performance.

\bibliographystyle{IEEEtran}
\bibliography{bib_test}

\end{document}